\documentclass[10pt,twocolumn,letterpaper]{article}

\usepackage[pagenumbers]{cvpr}      
\definecolor{cvprblue}{rgb}{0.21,0.49,0.74}
\usepackage[pagebackref,breaklinks,colorlinks,allcolors=cvprblue]{hyperref}
\usepackage{cuted}    
\usepackage{caption}  

\def\confName{CVPR}
\def\confYear{2027}

\title{ActiveBridge: Scaling Active Priors to Robots\\through Egocentric Perception-Interaction Modeling}

\author{%
Zixiao Gu$^{1,2,3,*}$,
Qiren Qin$^{2,4,*}$,
Yu He$^{2,3,4}$,
Zhengyang Liu$^{2,3,4}$,
Da Huang$^{1,2,3}$,\\
Hongquan Zhang$^{2,3,5}$,
Simo Wu$^{6}$,
Peng Zhou$^{6}$,
Guangnan Ye$^{3,4}$,
Yunbo Wang$^{1,\dagger}$,
Yanwei Fu$^{3,4,\dagger}$\\[4pt]
\resizebox{\dimexpr\textwidth-2\tabcolsep\relax}{!}{%
  \normalsize
  $^{1}$ Shanghai Jiao Tong University \quad
  $^{2}$ ActiMind \quad
  $^{3}$ Shanghai Innovation Institution \quad
  $^{4}$ Fudan University \quad
  $^{5}$ East China Normal University \quad
  $^{6}$ Luxi Tech%
}\\[3pt]
{\small
$^{*}$ Equal contribution. \quad
$^{\dagger}$ Corresponding authors.}
}

\begin{document}
\maketitle
\preCutedStrip{\vspace{-10pt}}
\begin{strip}
    \centering
    \includegraphics[width=\linewidth]{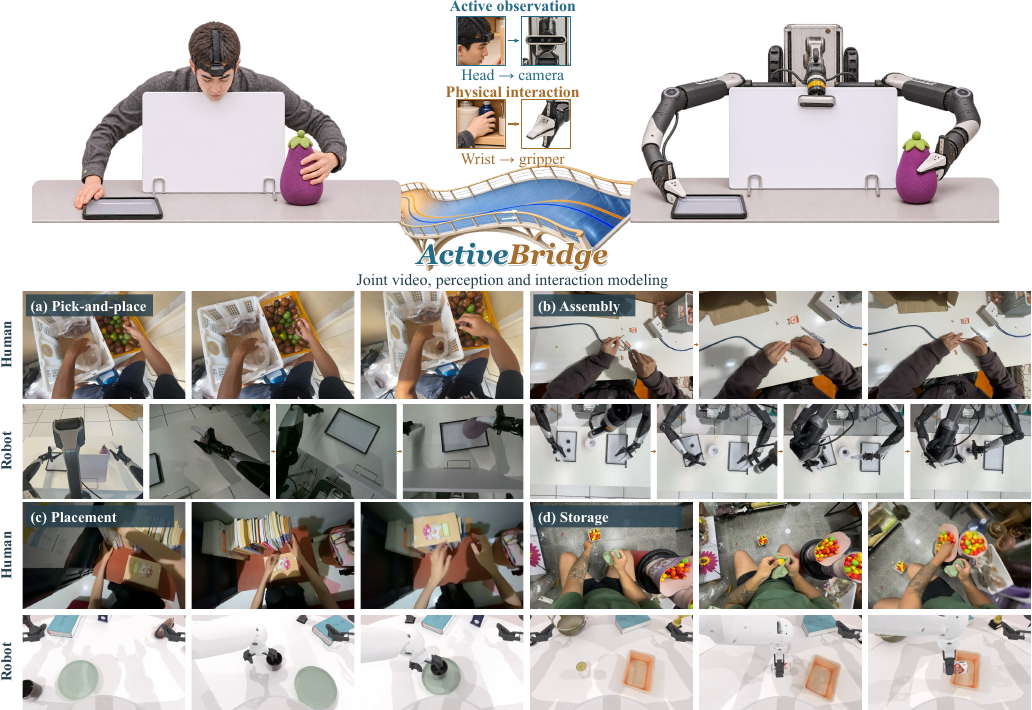}
    \captionof{figure}{\textbf{ActiveBridge learns human active priors from egocentric videos and transfers them to robots.}
    \textcolor[HTML]{246B85}{Top:} Head-wrist action trajectories from egocentric videos form a 21D human action representation. Jointly predicting future video and human actions helps the model learn active perception and physical interaction.
    \textcolor[HTML]{A66E2E}{Bottom:} Human training and robot deployment examples span real-world (a) pick-and-place and (b) assembly tasks, and simulation (c) placement and (d) storage tasks.}
    \label{fig:transfer}
\end{strip}

\begin{abstract}
Robotics foundation models are now widely used for manipulation, and egocentric videos offer rich data for scaling. However, existing methods overlook the human active priors captured in egocentric videos, which limits performance in unstructured scenes. They act on given observations instead of actively seeking information as humans do. We propose \textbf{ActiveBridge}, a training framework that learns these human active priors and transfers them to robots through world-action modeling. We design a robot that actively controls its viewpoint through head actions to acquire the visual information needed for manipulation using a single head camera, without wrist cameras. During human training, an instruction-conditioned model jointly predicts future egocentric videos and synchronized head-wrist action trajectories, with head actions providing explicit supervision for active perception. Experiments show that ActiveBridge outperforms baselines on active manipulation tasks, while ablations demonstrate the benefit of joint head-wrist action supervision.
\end{abstract}

\section{Introduction}
\label{sec:intro}

Robotics foundation models have expanded from multitask vision policies such as RT-1~\cite{brohan2022rt} to vision-language-action (VLA) models that adapt pretrained vision-language models for robot control~\cite{brohan2023rt,liu2025trivla}. Recent world-action models (WAMs) couple future prediction with action generation, drawing on pretrained video models to learn visual dynamics~\cite{lu2026world,li2026causal,ye2026world}. Under partial observability, manipulation also requires active perception: acquiring missing information through actions such as changing the viewpoint to locate an occluded object (\cref{fig:transfer}).

Data collection has also expanded beyond robot-specific demonstrations, with Open X-Embodiment pooling robot experience across embodiments~\cite{o2024open}. The Universal Manipulation Interface (UMI) enables robot-free collection with handheld grippers and a policy interface designed for transfer across robot platforms~\cite{chi2024universal}. Human egocentric video further extends the available experience to everyday interactions across diverse objects and environments~\cite{grauman2022ego4d,punamiya2026egoverse,lu2026world}. Recent methods incorporate human action supervision into WAMs~\cite{li2026egowam,sun2026ego4wam} and train policies that coordinate viewpoint changes with manipulation~\cite{li2026act}.

Egocentric video captures how people actively gather visual information, use it to guide physical interaction, and respond to the resulting visual changes~\cite{lin2026activemimic}. This raises our first question: \emph{how can egocentric video unlock active perception and interaction capabilities in existing world-action models?}
More specifically, the wrist-only action supervision used in previous work leaves active perception implicit~\cite{li2026egowam,sun2026ego4wam}. Transferring human active priors to robots also calls for closer alignment between human perception and robot sensing. Humans rely on head-centered vision for active perception, whereas many manipulation systems combine wrist cameras with a head or external camera~\cite{black2024pi_0,xiong2025vision}. Wrist-camera views are constrained by arm motion, and the additional hardware increases costs and collision risks. This raises our second question: \emph{how can active perception be explicitly supervised, and how can robot sensing be better aligned with human perception?}

We introduce \textbf{ActiveBridge}, a training framework that uses egocentric videos to train existing WAMs for coordinated active perception and physical interaction. Conditioned on instructions and observations, the model jointly predicts future egocentric videos and synchronized human action trajectories. By jointly modeling active perception and physical interaction, the model learns human priors that can be adapted to each robot embodiment. To provide explicit supervision for active perception, we include head actions as prediction targets alongside wrist trajectories and future video. The synchronized targets link head movements to the accompanying physical interactions and visual observations. To better align robot sensing with human perception, we design a platform with a single movable head camera and no wrist cameras. On this platform, the policy coordinates active perception and physical interaction by jointly controlling the head and arms. This design supports active manipulation while reducing sensor costs and collision risks associated with wrist-mounted hardware.

Our experiments validate the advantages of ActiveBridge. On the TAVIS-Head benchmark, which focuses on active manipulation, ActiveBridge outperforms baselines and the wrist-only variant across all six evaluation conditions. The in-distribution success rate reaches 77.1\% on Reachy-2 and 63.8\% on GR1T2. Under spatial distribution shifts, gains over direct adaptation reach 5.7\% and 3.4\% on Reachy-2 and GR1T2, respectively. Real-world evaluations further demonstrate the effectiveness of ActiveBridge across diverse manipulation tasks and the feasibility of our robot platform with a single movable head camera and no wrist cameras. The ablation study supports the benefit of the joint human video-action objective and explicit head-action supervision.

Our contributions are as follows:
\begin{itemize}
    \item \textbf{Human Active Prior Scaling from Egocentric Videos.} We explore how embodied models can learn from egocentric videos to actively acquire task-relevant perception and use it to guide physical interaction in unstructured environments, including those with occlusions.
    \item \textbf{A Novel ActiveBridge Framework and Active Robot Platform.} We propose ActiveBridge, which explicitly supervises active perception by predicting head actions. To align robot sensing with human perception, we design a platform with a single head camera.
    \item \textbf{State-of-the-Art Performance and Active Manipulation Improvement.} Experiments in simulation and the real world show that ActiveBridge outperforms baselines on active manipulation tasks. Ablations further demonstrate the performance improvements from ActiveBridge.
\end{itemize}
\section{Related Work}
\label{sec:related}

\noindent \textbf{World-action models.}
World-action models link control to future visual prediction. Early approaches use generated videos or visual subgoals to guide separate controllers~\cite{du2023learning,black2024zero}, or learn action-conditioned simulators~\cite{yang2023learning}. Subsequent methods integrate future prediction with action learning~\cite{wu2024unleashing,tian2025predictive}, jointly predict videos and actions using pretrained generators~\cite{shen2026videovla,ye2026world}, and develop causal video-action modeling for closed-loop control~\cite{li2026causal}. Further work examines value-guided planning~\cite{kim2026cosmos} as well as architectures, information flow, and training mixtures~\cite{wang2026openwam}.
Egocentric WAMs study prediction targets, data alignment, action supervision, and human--robot integration~\cite{li2026egowam,sun2026ego4wam,wu2026atomego}. AnyWorld factorizes action, camera, and embodiment controls to generate robot-domain experience~\cite{chen2026anyworld}. ActiveBridge extends wrist-supervised human WAM training with explicit head-action targets to jointly learn active perception, physical interaction, and their future visual consequences.

\noindent \textbf{Egocentric training.}
Egocentric videos support representation and reward learning~\cite{nair2022r3m,ma2022vip}, affordance extraction~\cite{bahl2023affordances}, and latent planning for control~\cite{wang2023mimicplay}. EgoVerse expands this resource with diverse demonstrations and shared transfer protocols~\cite{punamiya2026egoverse}. EgoMimic and HAT learn from human motion through joint human--robot training and shared state-action representations~\cite{kareer2025egomimic,qiu2025humanoid}, while EgoVLA predicts wrist and hand actions before retargeting and robot fine-tuning~\cite{yang2025egovla}.
Observation behavior also supplies supervision. CoMe-VLA incorporates visual and proprioceptive memory with subtask-transition supervision, while ActiveScale combines history, camera-pose targets, and mixed human--robot training~\cite{li2026act,zhou2026activescale}. ActiveBridge learns human active priors from egocentric demonstrations by jointly predicting future video and synchronized head-wrist trajectories. Head motion provides explicit supervision for viewpoint acquisition.

\noindent \textbf{Active perception.}
Active perception acquires information through sensing actions~\cite{bajcsy1988active}, while interactive perception also uses physical interaction~\cite{bohg2017interactive}. Learning to Look Around learns view-selection policies from uncertainty-reduction rewards~\cite{jayaraman2018learning}. For manipulation, AV-ALOHA uses an independently controlled camera arm, SaPaVe adopts separate camera and manipulation action representations~\cite{chuang2025active,liu2026sapave}, and ActiveVLA selects informative views and refines task-relevant 3D regions~\cite{liu2026activevla}. ActiveWAM focuses on retaining and acquiring evidence during robot control~\cite{wu2026activewam}. ActiveMimic learns active perception and manipulation by pretraining a VLA on synchronized camera and wrist trajectories recovered from egocentric human videos~\cite{lin2026activemimic}. AWAM predicts task-conditioned virtual camera trajectories and uses a geometry-conditioned world model to construct visual memory for action prediction~\cite{geng2026awam}. ActiveAlign adapts manipulation demonstrations to the states reached after viewpoint selection through closed-loop alignment and successful replay, then fine-tunes a VLA on these continuations~\cite{geng2026look}. TAVIS evaluates information acquisition and viewpoint control under distribution shifts~\cite{spigler2026tavis}. ActiveBridge learns observation–interaction priors from human egocentric demonstrations. To align robot sensing with these human priors, our robot platform uses a single movable head camera that provides shared visual feedback for joint head-and-arm control, without wrist cameras.
\section{ActiveBridge}
\label{sec:method}
\begin{figure*}[t]
\centering
\includegraphics[width=\linewidth]{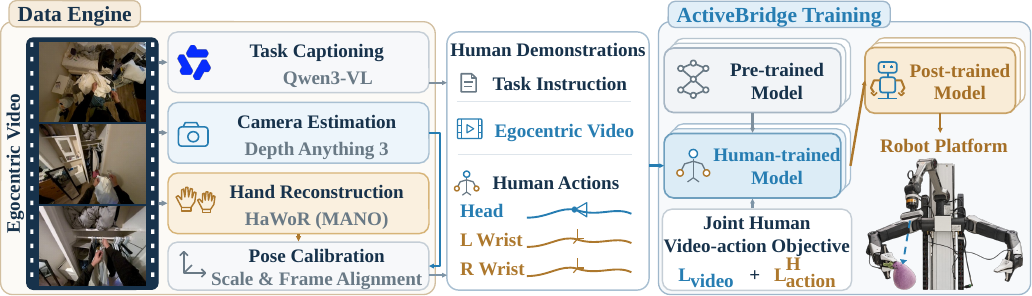}
\caption{\textbf{Overview of ActiveBridge.}
\emph{Left (Data Engine):} Task captioning, camera estimation, hand reconstruction, and pose calibration yield task instructions and synchronized head-wrist action trajectories in the same coordinate system.
\emph{Center (Human Demonstrations):} Each demonstration pairs a task instruction and egocentric video with 21D actions: the 3D position and 4D scalar-first quaternion for the head and each wrist.
\emph{Right (ActiveBridge Training):} A pretrained model learns from human demonstrations through a joint human video-action objective. The human-trained model then undergoes native robot post-training for deployment on our platform, which uses only a head camera to better align robot sensing with human perception.}
\label{fig:method}
\end{figure*}

ActiveBridge learns human active priors from egocentric demonstrations: how people acquire task-relevant perception and use it to guide physical interaction (\cref{fig:method}). Given a task instruction and observed frames, the model jointly predicts future egocentric video and synchronized head-wrist action trajectories. These targets connect actions for active perception and physical interaction to the visual observations that follow. We adapt these priors to coordinated head and arm control on our robot platform, which uses a single movable head camera and no wrist cameras.

\noindent \textbf{Overview.}
Our data engine pairs egocentric videos with task instructions and synchronized head-wrist trajectories to construct human demonstrations. During ActiveBridge training, the model first learns human active priors from these demonstrations through joint video-action prediction and then adapts to our platform's sensing and control interface through robot post-training. Let $\mathcal D_H$ denote human demonstrations and $\mathcal D_r$ denote demonstrations collected on our robot platform. Starting from a pretrained world-action model $\theta_0$, human training produces $\theta_H$, which initializes robot post-training to obtain $\theta_r$:
\begin{equation}
    \theta_0
    \xrightarrow{\mathcal D_H}\theta_H
    \xrightarrow{\mathcal D_r}\theta_r
\end{equation}

Here, $\theta_H$ transfers the human active prior, while $\mathcal D_r$ grounds it in our platform's observations and executable controls.

\subsection{Data Engine}
\label{sec:data_engine}

We construct human demonstrations from 387 hours of egocentric video selected from EgoVerse~\cite{punamiya2026egoverse}. The data engine associates each video with a task instruction and recovers synchronized head and wrist trajectories in the same coordinate system.

\noindent \textbf{Task captioning.} We use Qwen3-VL-8B-Instruct~\cite{bai2025qwen3} to caption the videos and obtain task instructions. Each instruction describes the recorded interaction in the form of an imperative sentence.

\noindent \textbf{Camera estimation.} We use Depth Anything 3~\cite{lin2025depth} to estimate metric-scale camera trajectories and use the camera poses as head poses. Camera translations are expressed in meters. Let $\bar T_t^{\mathrm{head}}\in SE(3)$ map coordinates from the camera frame at time $t$ to a common world frame. We obtain this camera-to-world pose by inverting the world-to-camera extrinsics.

\noindent \textbf{Wrist estimation.} We use HaWoR~\cite{zhang2025hawor} to reconstruct hand motion with the MANO model~\cite{romero2022embodied}. We extract the wrist position from the MANO root joint and use the predicted root rotation as its orientation. Let $\hat T_{t}^{b}\in SE(3)$ denote this wrist-to-camera pose for $b\in\{\mathrm{left},\mathrm{right}\}$, with translation expressed in meters.

\noindent \textbf{Pose calibration.} To express the wrist poses in the same coordinate system as the head pose, we combine the wrist poses with the metric head camera trajectory:
\begin{equation}
\bar T_t^{b}
=
\bar T_t^{\mathrm{head}}\hat T_t^{b},
\qquad b\in\{\mathrm{left},\mathrm{right}\}.
\label{eq:wrist_world}
\end{equation}
Here, $\bar T_t^{b}$ expresses a wrist pose in the same world frame as the head pose. This composition accounts for camera motion when expressing the reconstructed wrist trajectories.

\noindent \textbf{First-frame alignment.} We express the head and both wrist trajectories in a coordinate frame defined by the initial head pose. Let $(\bar R_t^d,\bar p_t^d)$ be the rotation and translation of $\bar T_t^d$, for $d\in\{\mathrm{head},\mathrm{left},\mathrm{right}\}$. The aligned position and orientation are
\begin{equation}
\begin{aligned}
p_t^d
&=(\bar R_1^{\mathrm{head}})^\top
  (\bar p_t^d-\bar p_1^{\mathrm{head}}),\\
R_t^d
&=(\bar R_1^{\mathrm{head}})^\top\bar R_t^d.
\end{aligned}
\label{eq:first_frame_alignment}
\end{equation}
The same reference rotation and translation apply to every time step and all three trajectories. Thus, the initial head pose has zero position and identity rotation, while the wrists retain their positions relative to that head pose. We synchronize these trajectories with the video to form the human demonstrations.

\subsection{Human Demonstrations}
\label{sec:human_demonstrations}
\label{sec:human_training}

The data engine produces $\mathcal D_H=\{(c,V^H,A^H)\}$. Each triplet combines a task instruction, an egocentric visual sequence, and synchronized head-wrist poses from the same demonstration.

\noindent \textbf{Task instruction.} The instruction $c$ specifies the goal of the recorded interaction. It conditions both video and action prediction, so the two modalities describe the same task.

\noindent \textbf{Egocentric video.} The sequence $V^H=(v_1^H,\ldots,v_T^H)$ contains $T$ temporally ordered RGB frames from the human head camera. Observed frames provide the visual context for prediction, while future frames supervise the visual observations that follow head and wrist action. Each frame $v_t^H$ is paired with the corresponding pose target $a_t^H$.

\noindent \textbf{Head-wrist actions.} The action sequence $A^H=(a_1^H,\ldots,a_T^H)$ contains the aligned head and wrist poses. At each time step $t$, the action target is
\begin{equation}
a_t^H=
\left[
p_t^{\mathrm{head}},q_t^{\mathrm{head}},
p_t^{\mathrm{left}},q_t^{\mathrm{left}},
p_t^{\mathrm{right}},q_t^{\mathrm{right}}
\right]\in\mathbb R^{21}.
\end{equation}
The labels $\{\mathrm{head},\mathrm{left},\mathrm{right}\}$ refer to the head, left wrist, and right wrist, respectively. Each pose comprises a 3D position $p_t^d$ in meters and a scalar-first orientation quaternion $q_t^d=[q_w,q_x,q_y,q_z]$, giving a 7D representation. Together, the three poses constitute a 21D human head-wrist action representation. The quaternion represents the aligned rotation $R_t^d$ in \cref{eq:first_frame_alignment}.

\subsection{ActiveBridge Training}
\label{sec:activebridge_training}

\noindent \textbf{Joint human video-action objective.} Conditioned on the instruction $c$ and observed egocentric video, the model jointly predicts future video and synchronized head-wrist actions:
\begin{equation}
\begin{aligned}
\mathcal L_H
&=\mathcal L_{\mathrm{video}}+\mathcal L_{\mathrm{action}}^H,\\
\mathcal L_{\mathrm{action}}^H
&=\lambda\mathcal L_{\mathrm{head}}^H+\mathcal L_{\mathrm{wrist}}^H.
\end{aligned}
\label{eq:human_objective}
\end{equation}
Here, $\mathcal L_{\mathrm{head}}^H$ and $\mathcal L_{\mathrm{wrist}}^H$ denote the contributions of the head and both wrist trajectories to the original action loss, respectively, while $\mathcal L_{\mathrm{video}}$ supervises future egocentric observations. $\lambda$ is the weight of head-action supervision.

We retain the flow-matching objective of diffusion transformers. For a video latent or preprocessed action target $x$, we form $x_\sigma=(1-\sigma)x+\sigma\epsilon$ at noise level $\sigma\in[0,1]$, with $\epsilon\sim\mathcal N(0,I)$. Given the predicted flow velocity $\hat u_x$, we define its residual as $e_x=\hat u_x-(\epsilon-x)$. Suppressing time indices, let $z$ denote video latents and $a^H,a^r$ denote the human and robot actions. The video loss is
\begin{equation}
\mathcal L_{\mathrm{video}}
=\mathbb E\!\left[\frac{w_v}{D_v}\|e_z\|_2^2\right],
\label{eq:video_flow_loss}
\end{equation}
where $D_v$ is the number of supervised latent elements. The human action residual is partitioned into one head block and two wrist blocks:
\begin{equation}
\begin{aligned}
\mathcal L_{\mathrm{head}}^H
&=\mathbb E\!\left[\frac{w_a^H}{21}\|e_{a^H}^{\mathrm{head}}\|_2^2\right],\\
\mathcal L_{\mathrm{wrist}}^H
&=\mathbb E\!\left[\frac{w_a^H}{21}
 \sum_{b\in\{\mathrm{left},\mathrm{right}\}}\|e_{a^H}^{b}\|_2^2\right].
\end{aligned}
\label{eq:head_wrist_flow_losses}
\end{equation}
Here, $w_v$ and $w_a^H,w_a^r$ are the inherited noise-level weights, and the expectations average over demonstrations, supervised time steps, and noise samples. Both action terms use the same 21D normalization.

The head term provides explicit head-action supervision for active perception, and the wrist term supervises physical interaction. Their synchronization with future video links both kinds of motion to the visual observations that follow under the same task instruction. Optimizing this objective from $\theta_0$ yields the human-trained model $\theta_H$, which initializes robot post-training.

\noindent \textbf{Robot platform and post-training.} Motivated by how humans actively perceive environments during interaction, our platform uses a single movable head camera and no wrist cameras. Its native action space $\mathcal A_r$ comprises head, arm, and gripper commands for viewpoint control and manipulation. We adapt the human prior using robot demonstrations
\begin{equation}
\mathcal D_r=
\{(c,V^r,A^r)\},
\end{equation}
where $V^r$ contains head-camera video and $A^r$ contains the corresponding native action trajectories in $\mathcal A_r$. Initialized from $\theta_H$, the model is conditioned on the task instruction $c$ and observed head-camera frames and optimized with the robot video-action objective:
\begin{equation}
\begin{aligned}
\mathcal L_r
&=\mathcal L_{\mathrm{video}}+\mathcal L_{\mathrm{action}}^r,\\
\mathcal L_{\mathrm{action}}^r
&=\mathbb E\!\left[\frac{w_a^r}{D_a^r}\|e_{a^r}\|_2^2\right],
\end{aligned}
\label{eq:robot_objective}
\end{equation}
where $D_a^r$ is the native action dimension. These expressions are written for complete targets; masking and valid-target normalization follow the base WAM.

Here, $\mathcal L_{\mathrm{video}}$ in \cref{eq:robot_objective} supervises future head-camera observations, while $\mathcal L_{\mathrm{action}}^r$ supervises the platform's native head, arm, and gripper commands. Initialization transfers the human active prior, while these robot-specific supervision targets adapt the model to our platform's visual observations and executable controls. Robot post-training yields the adapted model $\theta_r$.

\noindent \textbf{Deployment.}
At time $t$, the adapted predictor $F_{\theta_r}$ maps the instruction $c$ and head-camera observation context $h_t^r$ to future video and a native action sequence:
\begin{equation}
\left(\widehat V^r,\widehat A^r\right)
=
F_{\theta_r}(c,h_t^r).
\end{equation}
$\widehat V^r$ and $\widehat A^r$ denote the predicted future head-camera video and actions, respectively. The robot executes the predicted head, arm, and gripper commands and then updates its observation context with newly acquired head-camera frames. Subsequent predictions use these observations to coordinate active perception and physical interaction.
\section{Experiments}
\label{sec:exp}

Our experiments address three questions about learning human active priors: \textbf{Q1:} (TAVIS-Head Benchmark and Real-Robot Analysis) Does human training improve robot performance on tasks that require active perception and interaction? \textbf{Q2:} (Ablation Study) Does explicit head-action supervision improve model performance? \textbf{Q3:} (RoboTwin 2.0 Benchmark) Does the same human active prior also support general manipulation?

\begin{table*}[t]
\centering
\caption{\textbf{Results on the TAVIS-Head Benchmark.}
Results use native headcam observations, comprising the head camera and both wrist cameras, with success rate (\%, $\uparrow$) as the metric.
The Tasks column indicates the number of evaluated tasks out of five. Success rates are averaged over three tasks for Diffusion Policy and all five tasks for the other methods.
ID/S/P denote ID, spatial OOD, and initial-pose OOD, respectively.
\colorbox{red!50}{Red}, \colorbox{orange!50}{orange}, and \colorbox{yellow!50}{yellow} cells mark the best, second-best, and third-best methods. ActiveBridge achieves the highest success rates among methods evaluated on all five tasks in each setting.}
\label{tab:tavis}
\setlength{\tabcolsep}{12.0pt}
\begin{tabular}{lcrrrrrr}
\toprule
& & \multicolumn{3}{c}{Reachy-2 headcam} & \multicolumn{3}{c}{GR1T2 headcam} \\
\cmidrule(lr){3-5}\cmidrule(lr){6-8}
Method & Tasks & ID$\uparrow$ & S$\uparrow$ & P$\uparrow$ & ID$\uparrow$ & S$\uparrow$ & P$\uparrow$ \\
\midrule
\multicolumn{8}{l}{\small\itshape\color{black!60}VLA and policy baselines} \\
Diffusion Policy (single)~\cite{chi2023diffusion} & 3/5 & 46.9 & 26.4 & 22.2 & 54.2 & 33.3 & 24.0 \\
$\pi_0$ (single)~\cite{black2024pi_0} & 5/5 & 39.6 & 24.2 & 22.9 & 25.8 & 9.6 & 8.8 \\
$\pi_0$ (multi)~\cite{black2024pi_0} & 5/5 & 38.8 & 22.1 & 11.2 & 47.1 & 27.9 & 1.9 \\
$\pi_{0.5}$~\cite{intelligence2025pi05} & 5/5 & 44.0 & 27.0 & 16.3 & 51.0 & 28.7 & 7.7 \\
\midrule
\multicolumn{8}{l}{\small\itshape\color{black!60}WAM baselines} \\
Fast-WAM~\cite{yuan2026fast} & 5/5 & 41.0 & 27.0 & 17.0 & 47.0 & 30.0 & 6.0 \\
EasyWAM-Unified~\cite{easywam2026} & 5/5 & 46.0 & 27.3 & 20.7 & 51.3 & 30.0 & 11.7 \\
ActiveWAM~\cite{wu2026activewam} & 5/5 & 52.7 & 44.0 & 31.3 & \colorbox{orange!50}{63.7} & \colorbox{orange!50}{47.0} & 24.7 \\
\midrule
\multicolumn{8}{l}{\small\itshape\color{black!60}Ablations} \\
LingBot-VA~\cite{li2026causal} & 5/5 & \colorbox{orange!50}{76.8} & \colorbox{yellow!50}{50.2} & \colorbox{yellow!50}{72.5} & 61.8 & 44.5 & \colorbox{yellow!50}{56.7} \\
ActiveBridge (Wrist-only) & 5/5 & \colorbox{yellow!50}{74.7} & \colorbox{orange!50}{54.3} & \colorbox{orange!50}{74.8} & \colorbox{yellow!50}{63.1} & \colorbox{yellow!50}{46.6} & \colorbox{orange!50}{57.3} \\
\textbf{ActiveBridge (Ours)} & 5/5 & \colorbox{red!50}{77.1} & \colorbox{red!50}{55.9} & \colorbox{red!50}{76.1} & \colorbox{red!50}{63.8} & \colorbox{red!50}{47.9} & \colorbox{red!50}{58.1} \\
\bottomrule
\end{tabular}
\end{table*}

\begin{table}[t]
\centering
\caption{\textbf{Results on the RoboTwin 2.0 Benchmark.}
Results are reported as success rates (\%, $\uparrow$).
Avg. is computed from the unrounded scores as $(\mathrm{Clean}+\mathrm{Rand.})/2$, with equal weight assigned to each task.
Light blue rows highlight our methods.}
\label{tab:robotwin}
\setlength{\tabcolsep}{3.5pt}
\begin{tabular*}{\linewidth}{@{\extracolsep{\fill}}lrrr@{}}
\toprule
Method & Clean $\uparrow$ & Rand. $\uparrow$ & Avg. $\uparrow$ \\
\midrule
\multicolumn{4}{@{}l@{}}{\small\itshape\color{black!60}VLA baselines} \\
X-VLA~\cite{zheng2025xvla} & 72.9 & 72.8 & 72.85 \\
$\pi_0$~\cite{black2024pi_0} & 65.9 & 58.4 & 62.15 \\
$\pi_{0.5}$~\cite{intelligence2025pi05} & 82.7 & 76.8 & 79.75 \\
G0.5~\cite{liu2026g0} & 93.7 & 92.8 & 93.25 \\
Meituan-Robotics-0~\cite{team2026meituan} & 93.2 & 93.9 & 93.55 \\
InternVLA-A1.5~\cite{ma2026internvla} & 93.3 & 93.0 & 93.15 \\
ACE-Ego-0~\cite{li2026ace} & 91.1 & 90.6 & 90.87 \\
JoyAI-RA0.1~\cite{zhang2026joyai} & 90.5 & 89.3 & 89.88 \\
ABot-M0~\cite{yang2026abot} & 86.1 & 85.1 & 85.57 \\
VLAct-OFT~\cite{yang2026beyond} & 92.5 & 90.8 & 91.65 \\
TurboVLA~\cite{xie2026turbovla} & 88.0 & 88.1 & 88.06 \\
StarVLA-$\pi$~\cite{community2026starvla} & 88.1 & 88.8 & 88.45 \\
\begin{tabular}[c]{@{}l@{}}LingBot-VLA\\(w/o depth)~\cite{wu2026pragmatic}\end{tabular} & 86.5 & 85.3 & 85.92 \\
\midrule
\multicolumn{4}{@{}l@{}}{\small\itshape\color{black!60}WAM baselines} \\
Fast-WAM~\cite{yuan2026fast} & 91.9 & 91.8 & 91.83 \\
MotuBrain~\cite{team2026motubrain} & 95.8 & 96.1 & 95.95 \\
ImageWAM~\cite{zhang2026imagewam} & 93.2 & 93.6 & 93.38 \\
AHA-WAM~\cite{cai2026aha} & 93.4 & 92.2 & 92.80 \\
SelfWAM~\cite{pan2026selfwam} & 92.2 & 93.1 & 92.62 \\
Next Forcing~\cite{xu2026next} & 94.1 & 93.5 & 93.80 \\
\begin{tabular}[c]{@{}l@{}}TaskWAM\\(Fast-WAM)~\cite{li2026plausible}\end{tabular} & 92.7 & 93.1 & 92.88 \\
FlowWAM~\cite{chen2026flowwam} & 92.9 & 92.1 & 92.54 \\
LaWAM~\cite{chen2026lawam} & 92.6 & 89.8 & 91.22 \\
MVG-WAM~\cite{chen2026mvg} & 92.5 & 91.6 & 92.07 \\
Faster-WAM~\cite{ma2026faster} & 90.4 & 90.6 & 90.50 \\
Being-H0.7~\cite{luo2026being} & 90.2 & 89.6 & 89.90 \\
Light-WAM~\cite{li2026light} & 76.4 & 76.3 & 76.35 \\
X-WAM~\cite{guo2026unified} & 89.8 & 90.7 & 90.25 \\
Efficient-WAM~\cite{li2026efficient} & 86.7 & 85.7 & 86.20 \\
\midrule
\multicolumn{4}{@{}l@{}}{\small\itshape\color{black!60}Ablations} \\
LingBot-VA~\cite{li2026causal} & 91.6 & 90.8 & 91.20 \\
ActiveBridge (Wrist-only) & 91.8 & 89.8 & 90.80 \\
\rlap{\color{cvprblue!10}\rule[-\dp\strutbox]{\linewidth}{\dimexpr\ht\strutbox+\dp\strutbox\relax}}\textbf{ActiveBridge (Ours)} & 92.3 & 90.5 & 91.40 \\
\bottomrule
\end{tabular*}
\end{table}

\begin{table*}[t]
\centering
\caption{\textbf{Results on Real-World Tasks.}
Each entry reports the number of successful trials out of 10 attempts per task on our robot platform. The four task columns correspond to tasks (a)-(d) in \cref{fig:qualitative}.}
\label{tab:real_world}
\setlength{\tabcolsep}{5pt}
\renewcommand{\arraystretch}{1.12}
\begin{tabular*}{\textwidth}{@{\extracolsep{\fill}}lcccc@{}}
\toprule
Method & \shortstack{(a) Eggplant\\toy placement} & \shortstack{(b) Book\\organization} & \shortstack{(c) Dining\\table tidying} & \shortstack{(d) Bearing\\assembly} \\
\midrule
$\pi_0$ (VLA Baseline) & 10/10 & 10/10 & 6/10 & 1/10 \\
\midrule
\multicolumn{5}{@{}l@{}}{\small\itshape\color{black!60}Ablations}\\
LingBot-VA (WAM Baseline) & 9/10 & 7/10 & 1/10 & 0/10 \\
ActiveBridge (Wrist-only) & 10/10 & 9/10 & 3/10 & 0/10 \\
\rlap{\color{cvprblue!10}\rule[-\dp\strutbox]{\linewidth}{\dimexpr\ht\strutbox+\dp\strutbox\relax}}\textbf{ActiveBridge (Ours)} & 10/10 & 10/10 & 3/10 & 2/10 \\
\bottomrule
\end{tabular*}
\end{table*}

\begin{figure*}[t]
\centering
\includegraphics[width=\linewidth]{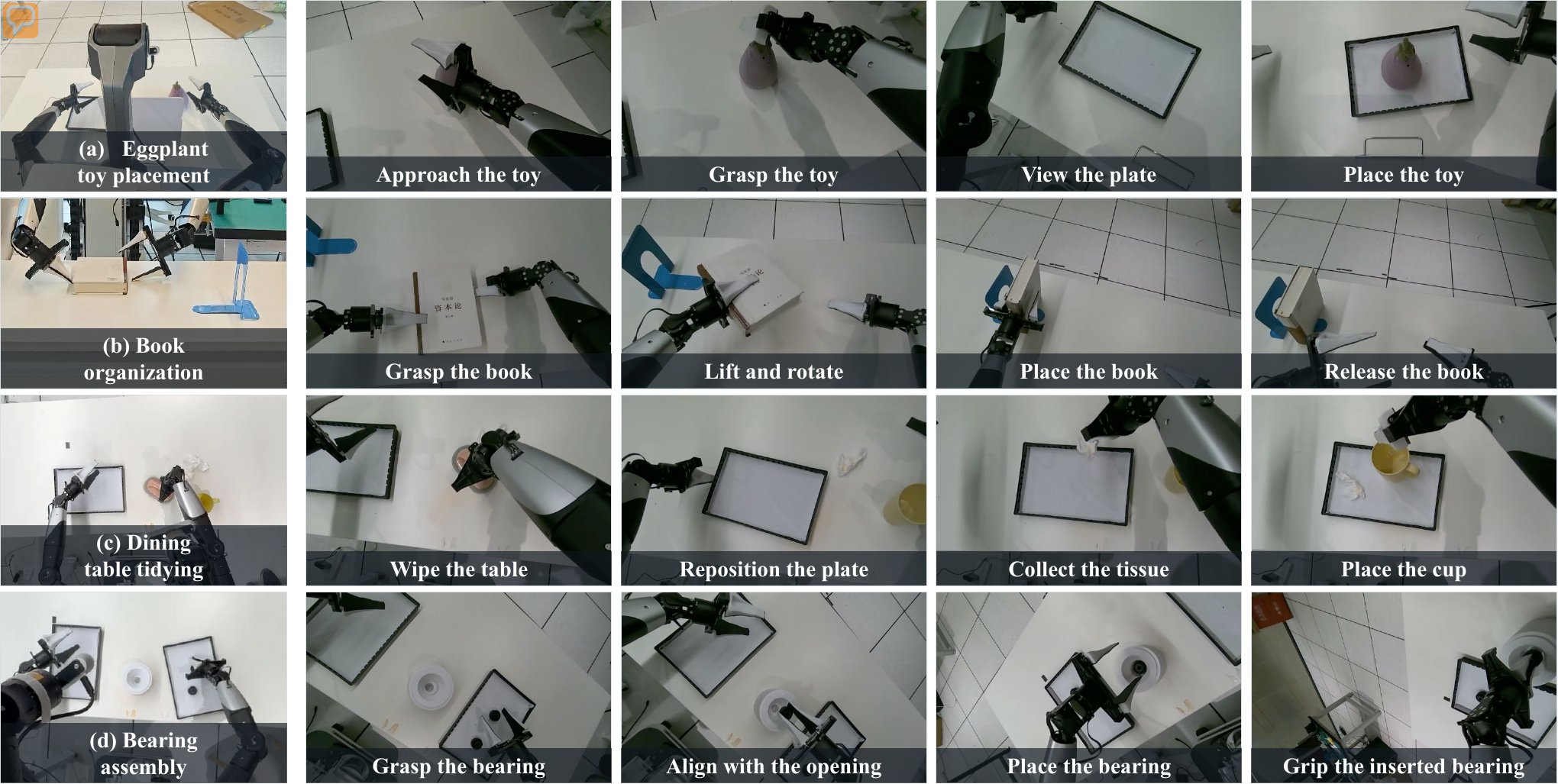}
\caption{\textbf{Progress across four real-world tasks: (a) eggplant plush placement, (b) book organization, (c) dining table tidying, and (d) bearing assembly.} Each task involves a sequence of manipulation primitives, \cref{tab:real_world} reports task success counts. The left column provides an exocentric scene overview, and each row of egocentric frames shows task milestones in temporal order.}
\label{fig:qualitative}
\end{figure*}

\subsection{Experimental Setup}

\noindent \textbf{Simulation Benchmarks.}
We use TAVIS-Head~\cite{spigler2026tavis} as our benchmark for active manipulation. TAVIS-Head includes five tasks for each of the Reachy-2 and GR1T2 robots: conditional-pick, wait-then-act, clutter-pick-cube, clutter-pick-lift, and multi-shelf-scan. We use native \emph{headcam} observations, including the head camera and both wrist cameras, with policy-controlled neck motion. ID (in-distribution) evaluation uses familiar task configurations, while spatial OOD (out-of-distribution) evaluation changes spatial configurations and initial-pose OOD evaluation changes initial pose configurations.
For general bimanual manipulation, we use RoboTwin 2.0~\cite{chen2025robotwin}, which evaluates 50 tasks under clean and randomized conditions. We focus the reference comparison on clean-plus-randomized training, with 50 clean and 500 randomized demonstrations per task~\cite{li2026causal}.

\noindent \textbf{Robot platform.}
Our robot platform uses two NERO arms for bimanual manipulation and a PiPER arm as an actuated neck, all from AgileX Robotics. The neck carries a single RealSense D455 camera and enables active viewpoint control alongside arm manipulation. We deliberately omit wrist cameras to reflect an egocentric observation setting and emphasize active perception through head motion. This minimal camera setup also reduces sensing hardware requirements and can lower the cost and complexity of real-world deployment. Omitting wrist cameras and their mounting brackets also removes potential collision hazards when manipulating objects in confined spaces.

\noindent \textbf{Real-world tasks.}
We evaluate four tasks: (a) eggplant toy placement, (b) book organization, (c) dining table tidying, and (d) bearing assembly. The first task places an eggplant plush toy on a plate. The book organization task involves arranging the book and the book stand. The dining table tidying task is a long-horizon task comprising different subtasks. The final task, bearing component assembly, is a precision machining task. Together, they cover object placement, scene organization, and part assembly with a single movable head camera. We collect 64 teleoperated demonstrations per task, jointly train on the combined dataset of 256 demonstrations, and conduct 10 evaluation trials per task.

\noindent \textbf{Baselines and ablations.}
For real-world tasks, we compare ActiveBridge with $\pi_0$~\cite{black2024pi_0}, LingBot-VA~\cite{li2026causal}, and our wrist-only variant. $\pi_0$ serves as a vision-language-action baseline, while LingBot-VA provides the world-action model baseline and the pretrained model used to initialize ActiveBridge. The wrist-only variant omits head-action targets during human training while retaining wrist-action supervision. The comparison with LingBot-VA evaluates the benefit of human training, while the comparison with the wrist-only variant evaluates the additional benefit of explicit head-action supervision. For simulation benchmarks, we include additional baselines (\cref{tab:tavis,tab:robotwin}) for a more comprehensive performance evaluation.

\subsection{Experimental Results}
\label{sec:active_results}

\noindent \textbf{Results on the TAVIS-Head Benchmark.}
ActiveBridge achieves the highest success rates in \cref{tab:tavis} across both robot embodiments and all three evaluation conditions under native headcam observations. It reaches 77.1\% ID success on Reachy-2 and 63.8\% on GR1T2. This lead persists under distribution shifts, with spatial OOD success rates of 55.9\% and 47.9\%, respectively, and 76.1\% and 58.1\% under initial-pose OOD. The gains over both VLA and world-action baselines suggest that jointly learning head-wrist actions and future observations provides a transferable prior for coordinating active perception with manipulation across embodiments and distribution shifts. Human training and explicit head-action supervision both contribute to these gains, and we analyze their effects in \cref{sec:head_ablation}.

\noindent \textbf{Results on the RoboTwin 2.0 Benchmark.}
ActiveBridge achieves success rates of 92.3\% in clean conditions and 90.5\% under randomization on RoboTwin 2.0 (\cref{tab:robotwin}). The mean success rate across the two conditions is 91.40\%, compared with 91.20\% of LingBot-VA. Relative to direct adaptation, success increases by 0.7\% in clean conditions and decreases by 0.3\% under randomization. We examine the effect of head-action supervision relative to wrist-only human training in \cref{sec:head_ablation}.

\noindent \textbf{Real-Robot Analysis.}
ActiveBridge improves over LingBot-VA on all four real-world tasks, raising the mean success rate from 42.5\% to 62.5\% (\cref{tab:real_world}). With ten trials per task, it matches $\pi_0$ on eggplant toy placement and book organization, achieving 10/10 successes on each. Bearing assembly yields 2/10 successes versus 1/10 for $\pi_0$, while dining table tidying remains challenging, with 3/10 successes compared with 6/10 for $\pi_0$. The rollouts in \cref{fig:qualitative} illustrate coordination between active perception and physical interaction using a single movable head camera. During toy placement, the view shifts from the object during grasping to the target plate for placement. The book organization rollout shows lifting, reorientation, and upright placement, while table tidying involves successive interactions with the wiping area, plate, tissue, and cup. Bearing assembly shows alignment with the opening, placement, and subsequent gripping of the inserted bearing. We analyze the contribution of explicit head-action supervision in \cref{sec:head_ablation}.

\subsection{Ablation Study}
\label{sec:head_ablation}

We compare ActiveBridge with two variants to evaluate the effects of the joint human video-action objective and explicit head-action supervision. LingBot-VA adapts the pretrained world-action model directly to each robot. The wrist-only variant retains joint prediction of future egocentric video and wrist trajectories during human training, while removing head-action targets.

\noindent \textbf{Effect of the joint human video-action objective.}
ActiveBridge outperforms LingBot-VA across both robot embodiments and all three TAVIS-Head evaluation conditions (\cref{tab:tavis}). In ID evaluation, ActiveBridge reaches 77.1\% success on Reachy-2 and 63.8\% on GR1T2. These scores exceed those of LingBot-VA by 0.3\% and 2.0\%, respectively. The largest gains occur under spatial OOD, where Reachy-2 success increases from 50.2\% to 55.9\%. On GR1T2, success increases from 44.5\% to 47.9\%. These improvements of 5.7\% and 3.4\% exceed the corresponding ID gains on both embodiments. Under initial-pose OOD, ActiveBridge also improves the success rate by 3.6\% on Reachy-2 and 1.4\% on GR1T2.

These improvements demonstrate the value of the joint human video-action objective during the additional human-training stage before robot adaptation. The gains under spatial and initial-pose shifts show that this benefit extends beyond familiar configurations. Together, the results support learning human active priors from synchronized egocentric video and head-wrist action trajectories as an effective initialization for active manipulation across embodiments.

\noindent \textbf{Effect of head-action supervision.}
ActiveBridge also outperforms its wrist-only variant in every TAVIS-Head condition, demonstrating the additional value of explicitly supervising head motion (\cref{tab:tavis}). On Reachy-2, ActiveBridge improves the success rate by 2.4\%, 1.6\%, and 1.3\% under ID, spatial OOD, and initial-pose OOD, respectively. On GR1T2, the corresponding gains are 0.7\%, 1.3\%, and 0.8\%. The improvements across both embodiments and all evaluation conditions support the benefit of including an explicit head-action target alongside wrist supervision during human training.

Although the wrist-only variant learns from egocentric video and wrist action trajectories, it does not fully exploit human active priors, as the head actions underlying viewpoint changes are not explicitly supervised. ActiveBridge additionally predicts the synchronized head-action trajectory to explicitly supervise the movement associated with visual observation (\cref{eq:human_objective}). On Reachy-2 ID, this addition raises success from 74.7\% to 77.1\%. The comparison shows that explicit head-action supervision adds value beyond the visual and wrist-action signals retained by the wrist-only variant, supporting the joint representation of perception and interaction.

The benefit of head-action supervision also extends to general bimanual manipulation on RoboTwin 2.0 (\cref{tab:robotwin}). ActiveBridge achieves 92.3\% success in clean conditions and 90.5\% under randomization. These scores exceed those of the wrist-only variant by 0.5\% and 0.7\%, respectively. The average across the two conditions increases from 90.80\% to 91.40\%, a gain of 0.60\%. These results demonstrate that the contribution of human head-action supervision transfers beyond benchmarks for active manipulation.
\section{Conclusion}
\label{sec:conclusion}

We present ActiveBridge, a training framework that learns human active priors from egocentric demonstrations and transfers them to robots through world-action modeling. By jointly predicting future video and synchronized head-wrist action trajectories, ActiveBridge explicitly supervises active perception alongside physical interaction. This formulation captures viewpoint motion and manipulation within a shared predictive framework, making their coordination an explicit part of human training. On TAVIS-Head, ActiveBridge outperforms baselines and two ablated variants across both embodiments and all evaluation conditions. Head-action supervision also improves general bimanual manipulation on RoboTwin 2.0 compared with wrist-only training. Real-world experiments demonstrate coordinated perception and interaction with a single movable head camera and no wrist cameras. Together, these results highlight the value of explicit head-action supervision for transferring human active priors across embodiments and manipulation settings. These findings show that learning how humans acquire visual information and act on it can improve robot manipulation.

{\small\bibliographystyle{ieeenat_fullname}\bibliography{main}}

@inproceedings{grauman2022ego4d,
  title={Ego4d: Around the world in 3,000 hours of egocentric video},
  author={Grauman, Kristen and Westbury, Andrew and Byrne, Eugene and Chavis, Zachary and Furnari, Antonino and Girdhar, Rohit and Hamburger, Jackson and Jiang, Hao and Liu, Miao and Liu, Xingyu and others},
  booktitle={Proceedings of the IEEE/CVF conference on computer vision and pattern recognition},
  pages={18995--19012},
  year={2022}
}

@article{li2026causal,
  title={Causal world modeling for robot control},
  author={Li, Lin and Zhang, Qihang and Luo, Yiming and Yang, Shuai and Wang, Ruilin and Han, Fei and Yu, Mingrui and Gao, Zelin and Xue, Nan and Zhu, Xing and others},
  journal={arXiv preprint arXiv:2601.21998},
  year={2026}
}

@inproceedings{li2026egowam,
  title={EgoWAM: World Action Models Beyond Pixels with In-the-Wild Egocentric Human Data},
  author={Li, Baoyu and Yin, Xinchen and Lin, Mengying and Zhang, Yixin and Xu, Danfei},
  booktitle={Robot World Models},
  year={2026}
}

@article{wu2026activewam,
  title={ActiveWAM: Evidence-Aware Active Vision for World-Action Models},
  author={Wu, Renjun and Ge, Luzhou and Li, Xuesong},
  journal={arXiv preprint arXiv:2610.01698},
  year={2026}
}

@article{spigler2026tavis,
  title={TAVIS: A Benchmark for Egocentric Active Vision and Anticipatory Gaze in Imitation Learning},
  author={Spigler, Giacomo},
  journal={arXiv preprint arXiv:2605.07943},
  year={2026}
}

@article{chen2025robotwin,
  title={Robotwin 2.0: A scalable data generator and benchmark with strong domain randomization for robust bimanual robotic manipulation},
  author={Chen, Tianxing and Chen, Zanxin and Chen, Baijun and Cai, Zijian and Liu, Yibin and Li, Zixuan and Liang, Qiwei and Lin, Xianliang and Ge, Yiheng and Gu, Zhenyu and others},
  journal={arXiv preprint arXiv:2506.18088},
  year={2025}
}

@inproceedings{kareer2025egomimic,
  title={Egomimic: Scaling imitation learning via egocentric video},
  author={Kareer, Simar and Patel, Dhruv and Punamiya, Ryan and Mathur, Pranay and Cheng, Shuo and Wang, Chen and Hoffman, Judy and Xu, Danfei},
  booktitle={2025 IEEE International Conference on Robotics and Automation (ICRA)},
  pages={13226--13233},
  year={2025},
  organization={IEEE}
}

@article{qiu2025humanoid,
  title={Humanoid policy\~{} human policy},
  author={Qiu, Ri-Zhao and Yang, Shiqi and Cheng, Xuxin and Chawla, Chaitanya and Li, Jialong and He, Tairan and Yan, Ge and Yoon, David J and Hoque, Ryan and Paulsen, Lars and others},
  journal={arXiv preprint arXiv:2503.13441},
  year={2025}
}

@article{li2026act,
  title={Act, Sense, Act: Learning Non-Markovian Active Perception Strategies from Large-Scale Egocentric Human Data},
  author={Li, Jialiang and Qiao, Yi and Guo, Yunhan and Chen, Changwen and Lian, Wenzhao},
  journal={arXiv preprint arXiv:2602.04600},
  year={2026}
}

@article{zhou2026activescale,
  title={ActiveScale: Scaling Active Perception for Robots across Model, Data, and Hardware},
  author={Zhou, Shuai and Pang, Kaisheng and Song, Wenxuan and Zhang, Wenjie and Zheng, Xinhu and Li, Haoang},
  journal={arXiv preprint arXiv:2609.18514},
  year={2026}
}

@article{shen2026videovla,
  title={Videovla: Video generators can be generalizable robot manipulators},
  author={Shen, Yichao and Wei, Fangyun and Du, Zhiying and Liang, Yaobo and Lu, Yan and Yang, Jiaolong and Zheng, Nanning and Guo, Baining},
  journal={Advances in neural information processing systems},
  volume={38},
  pages={95597--95621},
  year={2026}
}

@article{ye2026world,
  title={World action models are zero-shot policies},
  author={Ye, Seonghyeon and Ge, Yunhao and Zheng, Kaiyuan and Gao, Shenyuan and Yu, Sihyun and Kurian, George and Indupuru, Suneel and Tan, You Liang and Zhu, Chuning and Xiang, Jiannan and others},
  journal={arXiv preprint arXiv:2602.15922},
  year={2026}
}

@article{kim2026cosmos,
  title={Cosmos policy: Fine-tuning video models for visuomotor control and planning},
  author={Kim, Moo Jin and Gao, Yihuai and Lin, Tsung-Yi and Lin, Yen-Chen and Ge, Yunhao and Lam, Grace and Liang, Percy and Song, Shuran and Liu, Ming-Yu and Finn, Chelsea and others},
  journal={arXiv preprint arXiv:2601.16163},
  year={2026}
}

@article{wang2026openwam,
  title={Openwam: An open, modular exploration towards systematic world-action model pretraining},
  author={Wang, Yuran and Huang, Siqiao and Li, Mingleyang and Zhang, Chenhao and Liang, Jiaqi and Jin, Weiyang and Chen, Yue and Chi, Xuemin and Zhou, Donghao and Yu, Qize and others},
  journal={arXiv preprint arXiv:2609.07398},
  year={2026}
}

@article{sun2026ego4wam,
  title={Ego4WAM: What Matters When Scaling Egocentric Human Data for Robot Learning?},
  author={Sun, Zhihao and Liu, Liu and Wang, Xinjiang and Jiang, Haoyi and Feng, Wei and Zhang, Huiqiang and Jia, Xiaosong and Su, Zhizhong and Wu, Zuxuan},
  journal={arXiv preprint arXiv:2609.40341},
  year={2026}
}

@article{wu2026atomego,
  title={AtomEgo: Exploring Ego-Robot Integration for Embodied Foundation Model Pretraining},
  author={Wu, Di and Zheng, Dongchen and Sheng, Junhe and Wei, Zhongxing and Zhang, Songxin and Xie, Zejian and Sun, Xiaoquan and Zheng, Junyang and Song, Zhuoyang and Zhang, Jiaxing and others},
  journal={arXiv preprint arXiv:2609.21461},
  year={2026}
}

@inproceedings{chuang2025active,
  title={Active vision might be all you need: Exploring active vision in bimanual robotic manipulation},
  author={Chuang, Ian and Lee, Andrew and Gao, Dechen and Naddaf-Sh, M-Mahdi and Soltani, Iman},
  booktitle={2025 IEEE International Conference on Robotics and Automation (ICRA)},
  pages={7952--7959},
  year={2025},
  organization={IEEE}
}

@article{geng2026awam,
  title={AWAM: Active Trajectory-Conditioned View Imagination for Robust Robot Manipulation},
  author={Geng, Jiawei and Liu, Zhenyang and Wu, Yuzheng and Fu, Yanwei},
  journal={arXiv preprint arXiv:2610.23396},
  year={2026}
}

@article{geng2026look,
  title={Look Before You Act: Bridging Active Perception and Vision-Language-Action Models via Observation-Aligned Data Adaptation},
  author={Geng, Jiawei and Liu, Zhenyang and Wu, Yuzheng and Gu, Yongchong and Fu, Yanwei},
  journal={arXiv preprint arXiv:2610.23378},
  year={2026}
}

@article{lin2026activemimic,
  title={ActiveMimic: Egocentric Video Pretraining with Active Perception},
  author={Lin, Xingyao and Zhong, Guojin and Lu, Tianyi and Ye, Ziyi and Zhu, Yichen and Wu, Zuxuan and Jiang, Yu-Gang},
  journal={arXiv preprint arXiv:2606.06194},
  year={2026}
}

@article{liu2025trivla,
  title={TriVLA: A Triple-System-Based Unified Vision-Language-Action Model with Episodic World Modeling for General Robot Control},
  author={Liu, Zhenyang and Gu, Yongchong and Zheng, Sixiao and Fu, Yanwei and Xue, Xiangyang and Jiang, Yu-Gang},
  journal={arXiv preprint arXiv:2507.01424},
  year={2025}
}

@article{liu2026sapave,
  title={Sapave: Towards active perception and manipulation in vision-language-action models for robotics},
  author={Liu, Mengzhen and Zhou, Enshen and Chi, Cheng and Han, Yi and Rong, Shanyu and Chen, Liming and Wang, Pengwei and Wang, Zhongyuan and Zhang, Shanghang},
  journal={arXiv preprint arXiv:2603.12193},
  year={2026}
}

@article{chen2026anyworld,
  title={Anyworld: Factorized egocentric world models for cross-embodiment generalization},
  author={Chen, Cheng and Bai, Jerry and Wei, Jiacheng and Chen, Boyu and Zheng, Xiaoji and Wu, Fan and Yang, Minghao and Chen, Tianrun and Li, Ruibo and Yue, Xiaoyu and others},
  journal={arXiv preprint arXiv:2608.29242},
  year={2026}
}

@article{du2023learning,
  title={Learning universal policies via text-guided video generation},
  author={Du, Yilun and Yang, Sherry and Dai, Bo and Dai, Hanjun and Nachum, Ofir and Tenenbaum, Josh and Schuurmans, Dale and Abbeel, Pieter},
  journal={Advances in neural information processing systems},
  volume={36},
  pages={9156--9172},
  year={2023}
}

@article{yang2023learning,
  title={Learning interactive real-world simulators},
  author={Yang, Mengjiao and Du, Yilun and Ghasemipour, Kamyar and Tompson, Jonathan and Schuurmans, Dale and Abbeel, Pieter},
  journal={arXiv preprint arXiv:2310.06114},
  volume={1},
  number={2},
  pages={6},
  year={2023}
}

@inproceedings{wu2024unleashing,
  title={Unleashing large-scale video generative pre-training for visual robot manipulation},
  author={Wu, Hongtao and Jing, Ya and Cheang, Chilam and Chen, Guangzeng and Xu, Jiafeng and Li, Xinghang and Liu, Minghuan and Li, Hang and Kong, Tao},
  booktitle={International Conference on Learning Representations},
  volume={2024},
  pages={10641--10662},
  year={2024}
}

@inproceedings{tian2025predictive,
  title={Predictive inverse dynamics models are scalable learners for robotic manipulation},
  author={Tian, Yang and Yang, Sizhe and Zeng, Jia and Wang, Ping and Lin, Dahua and Dong, Hao and Pang, Jiangmiao},
  booktitle={International Conference on Learning Representations},
  volume={2025},
  pages={92033--92052},
  year={2025}
}

@inproceedings{black2024zero,
  title={Zero-shot robotic manipulation with pre-trained image-editing diffusion models},
  author={Black, Kevin and Nakamoto, Mitsuhiko and Atreya, Pranav and Walke, Homer and Finn, Chelsea and Kumar, Aviral and Levine, Sergey},
  booktitle={International Conference on Learning Representations},
  volume={2024},
  pages={33431--33452},
  year={2024}
}

@article{nair2022r3m,
  title={R3m: A universal visual representation for robot manipulation},
  author={Nair, Suraj and Rajeswaran, Aravind and Kumar, Vikash and Finn, Chelsea and Gupta, Abhinav},
  journal={arXiv preprint arXiv:2203.12601},
  year={2022}
}

@article{ma2022vip,
  title={Vip: Towards universal visual reward and representation via value-implicit pre-training},
  author={Ma, Yecheng Jason and Sodhani, Shagun and Jayaraman, Dinesh and Bastani, Osbert and Kumar, Vikash and Zhang, Amy},
  journal={arXiv preprint arXiv:2210.00030},
  year={2022}
}

@article{wang2023mimicplay,
  title={Mimicplay: Long-horizon imitation learning by watching human play},
  author={Wang, Chen and Fan, Linxi and Sun, Jiankai and Zhang, Ruohan and Fei-Fei, Li and Xu, Danfei and Zhu, Yuke and Anandkumar, Anima},
  journal={arXiv preprint arXiv:2302.12422},
  year={2023}
}

@inproceedings{bahl2023affordances,
  title={Affordances from human videos as a versatile representation for robotics},
  author={Bahl, Shikhar and Mendonca, Russell and Chen, Lili and Jain, Unnat and Pathak, Deepak},
  booktitle={2023 IEEE/CVF Conference on Computer Vision and Pattern Recognition (CVPR)},
  pages={01--13},
  year={2023},
  organization={IEEE}
}

@article{yang2025egovla,
  title={Egovla: Learning vision-language-action models from egocentric human videos},
  author={Yang, Ruihan and Yu, Qinxi and Wu, Yecheng and Yan, Rui and Li, Borui and Cheng, An-Chieh and Zou, Xueyan and Fang, Yunhao and Cheng, Xuxin and Qiu, Ri-Zhao and others},
  journal={arXiv preprint arXiv:2507.12440},
  year={2025}
}

@article{punamiya2026egoverse,
  title={Egoverse: An egocentric human dataset for robot learning from around the world},
  author={Punamiya, Ryan and Kareer, Simar and Liu, Zeyi and Citron, Josh and Qiu, Ri-Zhao and Cai, Xiongyi and Gavryushin, Alexey and Chen, Jiaqi and Liconti, Davide and Zhu, Lawrence Y and others},
  journal={arXiv preprint arXiv:2604.07607},
  year={2026}
}

@article{bajcsy1988active,
  title={Active perception},
  author={Bajcsy, Ruzena},
  journal={Proceedings of the IEEE},
  volume={76},
  number={8},
  pages={966--1005},
  year={1988},
  publisher={IEEE}
}

@article{bohg2017interactive,
  title={Interactive perception: Leveraging action in perception and perception in action},
  author={Bohg, Jeannette and Hausman, Karol and Sankaran, Bharath and Brock, Oliver and Kragic, Danica and Schaal, Stefan and Sukhatme, Gaurav S},
  journal={IEEE Transactions on Robotics},
  volume={33},
  number={6},
  pages={1273--1291},
  year={2017},
  publisher={IEEE}
}

@inproceedings{jayaraman2018learning,
  title={Learning to look around: Intelligently exploring unseen environments for unknown tasks},
  author={Jayaraman, Dinesh and Grauman, Kristen},
  booktitle={2018 IEEE/CVF Conference on Computer Vision and Pattern Recognition},
  pages={1238--1247},
  year={2018},
  organization={IEEE}
}

@article{liu2026activevla,
  title={ActiveVLA: Injecting active perception into vision-language-action models for precise 3D robotic manipulation},
  author={Liu, Zhenyang and Gu, Yongchong and Wang, Yikai and Xue, Xiangyang and Fu, Yanwei},
  journal={arXiv preprint arXiv:2601.08325},
  year={2026}
}

@article{bai2025qwen3,
  title={Qwen3-vl technical report},
  author={Bai, Shuai and Cai, Yuxuan and Chen, Ruizhe and Chen, Keqin and Chen, Xionghui and Cheng, Zesen and Deng, Lianghao and Ding, Wei and Gao, Chang and Ge, Chunjiang and others},
  journal={arXiv preprint arXiv:2511.21631},
  year={2025}
}

@article{lin2025depth,
  title={Depth anything 3: Recovering the visual space from any views},
  author={Lin, Haotong and Chen, Sili and Liew, Junhao and Chen, Donny Y and Li, Zhenyu and Shi, Guang and Feng, Jiashi and Kang, Bingyi},
  journal={arXiv preprint arXiv:2511.10647},
  year={2025}
}

@inproceedings{zhang2025hawor,
  title={Hawor: World-space hand motion reconstruction from egocentric videos},
  author={Zhang, Jinglei and Deng, Jiankang and Ma, Chao and Potamias, Rolandos Alexandros},
  booktitle={2025 IEEE/CVF Conference on Computer Vision and Pattern Recognition (CVPR)},
  pages={1805--1815},
  year={2025},
  organization={IEEE}
}

@article{romero2022embodied,
  title={Embodied hands: Modeling and capturing hands and bodies together},
  author={Romero, Javier and Tzionas, Dimitrios and Black, Michael J},
  journal={arXiv preprint arXiv:2201.02610},
  year={2022}
}

@article{black2024pi_0,
  title={$\pi_0$: A Vision-Language-Action Flow Model for General Robot Control},
  author={Black, Kevin and Brown, Noah and Driess, Danny and Esmail, Adnan and Equi, Michael and Finn, Chelsea and Fusai, Niccolo and Groom, Lachy and Hausman, Karol and Ichter, Brian and others},
  journal={arXiv preprint arXiv:2410.24164},
  year={2024}
}

@article{xiong2025vision,
  title={Vision in action: Learning active perception from human demonstrations},
  author={Xiong, Haoyu and Xu, Xiaomeng and Wu, Jimmy and Hou, Yifan and Bohg, Jeannette and Song, Shuran},
  journal={arXiv preprint arXiv:2506.15666},
  year={2025}
}

@article{brohan2022rt,
  title={Rt-1: Robotics transformer for real-world control at scale},
  author={Brohan, Anthony and Brown, Noah and Carbajal, Justice and Chebotar, Yevgen and Dabis, Joseph and Finn, Chelsea and Gopalakrishnan, Keerthana and Hausman, Karol and Herzog, Alex and Hsu, Jasmine and others},
  journal={arXiv preprint arXiv:2212.06817},
  year={2022}
}

@article{brohan2023rt,
  title={Rt-2: Vision-language-action models transfer web knowledge to robotic control},
  author={Brohan, Anthony and Brown, Noah and Carbajal, Justice and Chebotar, Yevgen and Chen, Xi and Choromanski, Krzysztof and Ding, Tianli and Driess, Danny and Dubey, Avinava and Finn, Chelsea and others},
  journal={arXiv preprint arXiv:2307.15818},
  year={2023}
}

@article{lu2026world,
  title={World-action models for robot learning and control: A survey},
  author={Lu, Zuxing and Zhai, Hongjia and Wang, Guanzhi and Zeng, Huajian and Yang, Jiaqi and Liu, Jingyu and Cheng, Lei and Zhang, Yuantai and Qiu, Yuheng and Cheng, Zezhou and others},
  journal={arXiv preprint arXiv:2609.16074},
  year={2026}
}

@inproceedings{o2024open,
  title={Open x-embodiment: Robotic learning datasets and rt-x models: Open x-embodiment collaboration 0},
  author={O’Neill, Abby and Rehman, Abdul and Maddukuri, Abhiram and Gupta, Abhishek and Padalkar, Abhishek and Lee, Abraham and Pooley, Acorn and Gupta, Agrim and Mandlekar, Ajay and Jain, Ajinkya and others},
  booktitle={2024 IEEE International Conference on Robotics and Automation (ICRA)},
  pages={6892--6903},
  year={2024},
  organization={IEEE}
}

@article{chi2024universal,
  title={Universal manipulation interface: In-the-wild robot teaching without in-the-wild robots},
  author={Chi, Cheng and Xu, Zhenjia and Pan, Chuer and Cousineau, Eric and Burchfiel, Benjamin and Feng, Siyuan and Tedrake, Russ and Song, Shuran},
  journal={arXiv preprint arXiv:2402.10329},
  year={2024}
}

@article{yuan2026fast,
  title={Fast-wam: Do world action models need test-time future imagination?},
  author={Yuan, Tianyuan and Dong, Zibin and Liu, Yicheng and Zhao, Hang},
  journal={arXiv preprint arXiv:2603.16666},
  year={2026}
}

@article{liu2026g0,
  title={G0. 5: One autoregressive stream for robot reasoning and action},
  author={Liu, Yicheng and Dong, Zibin and Ye, Baijun and Yuan, Tianyuan and Jiang, Tao and Yang, Anqi and Cao, Shicheng and Liu, Haonan and Sun, Yue and Guo, Zihan and others},
  journal={arXiv preprint arXiv:2608.11739},
  year={2026}
}

@article{team2026motubrain,
  title={Motubrain: An advanced world action model for robot control},
  author={Team, MotuBrain and Xiang, Chendong and Bao, Fan and Liu, Haitian and Tan, Hengkai and Bi, Hongzhe and Li, James and Liu, Jiabao and Pang, Jingrui and Jing, Kiro and others},
  journal={arXiv preprint arXiv:2604.27792},
  year={2026}
}

@article{chen2026mvg,
  title={MVG-WAM: Multiple View Geometry-Aware World-Action Modeling for Robotic Manipulation},
  author={Chen, Wenbo and Li, Tianfu and Xu, Haoxuan and Cao, Zhihao and Chen, Zhenghan and Zhu, Zhengming and Luo, Zizhou and Yang, Guosheng and Liu, Yuan and Wang, Lujia and others},
  journal={arXiv preprint arXiv:2609.37793},
  year={2026}
}

@article{team2026meituan,
  title={Meituan-Robotics-0: A Vision-Language-Action Foundation Model for Desktop Manipulation},
  author={Team, Meituan Robotics and Wang, Caicheng and Zhao, Chen and Jin, Chengpeng and Xu, Duo and Chen, Fengjiao and Wan, Guanglu and Yang, Heqing and Wang, Jia and Zhou, Jiacong and others},
  journal={arXiv preprint arXiv:2611.08577},
  year={2026}
}

@article{ma2026internvla,
  title={InternVLA-A1. 5: Unifying Understanding, Latent Foresight, and Action for Compositional Generalization},
  author={Ma, Haoxiang and Cai, Junhao and Xu, Xiaoxu and Li, Hao and Yang, Yuyin and Tian, Yang and Cao, Jiafei and Zhu, Hongrui and Qiu, Zherui and Yang, Yuqiang and others},
  journal={arXiv preprint arXiv:2607.04988},
  year={2026}
}

@article{zhang2026imagewam,
  title={ImageWAM: Do World Action Models Really Need Video Generation, or Just Image Editing?},
  author={Zhang, Yuyang and Zhang, Wenyao and Qi, Zekun and Zhang, He and Lin, Haitao and Zhang, Jingbo and Mu, Yao and Yang, Xiaokang and Zeng, Wenjun and Jin, Xin},
  journal={arXiv preprint arXiv:2606.19531},
  year={2026}
}

@article{cai2026aha,
  title={AHA-WAM: Asynchronous Horizon-Adaptive World-Action Modeling with Observation-Guided Context Routing},
  author={Cai, Jisong and Ling, Long and Chu, Shiwei and Liu, Zhongshan and Kang, Jiayue and Liang, Zhixuan and Xu, Wenjie and Mao, Yinan and Zhang, Weinan and Yang, Xiaokang and others},
  journal={arXiv preprint arXiv:2606.09811},
  year={2026}
}

@article{pan2026selfwam,
  title={Selfwam: A self-grounded unified world action model for fast robot control},
  author={Pan, Bikang and Liu, Fan and Lu, Haotao and Wang, Jingya and Shi, Ye},
  journal={arXiv preprint arXiv:2608.00725},
  year={2026}
}

@article{xu2026next,
  title={Next forcing: Causal world modeling with multi-chunk prediction},
  author={Xu, Gangwei and Zhang, Qihang and Zhou, Jiaming and Zhu, Xing and Shen, Yujun and Yang, Xin and Xu, Yinghao},
  journal={arXiv preprint arXiv:2606.11187},
  year={2026}
}

@article{li2026plausible,
  title={Plausible Future, Wrong Choice: Steering World-Action Models toward Intended Future},
  author={Li, Zhuohao and He, Jiawei and Yan, Yuxuan and Zhou, Lang and Li, Yinghao and Li, Yanrui and Zhang, Tianyu and Song, Yi-Yang and Hong, Fating and Zhang, Qing and others},
  journal={arXiv preprint arXiv:2610.28431},
  year={2026}
}

@article{chen2026flowwam,
  title={FlowWAM: Optical Flow as a Unified Action Representation for World Action Models},
  author={Chen, Yixiang and Li, Peiyan and Xu, Yuan and Ma, Qisen and Yang, Jiabing and Wang, Kai and Yang, Jianhua and An, Dong and Guan, He and Liu, Gaoteng and others},
  journal={arXiv preprint arXiv:2607.13017},
  year={2026}
}

@article{chen2026lawam,
  title={Lawam: Latent world action models for efficient dynamics-aware robot policies},
  author={Chen, Jialei and Wang, Kai and Chen, Kang and Chen, Shuaihang and Gao, Feng and Tang, Wenhao and Li, Zhiyuan and Liu, Weilin and Yao, Zhuyu and Li, Boxun and others},
  journal={arXiv preprint arXiv:2606.15768},
  year={2026}
}

@article{ma2026faster,
  title={Faster-WAM: Do World Action Models Need Deep Action Modules?},
  author={Ma, Liheng and Yang, Rui Heng and Zhang, Zhanguang and Clemente, Mateo and Hu, Ziwen and Cao, Tongtong and Zhang, Yingxue},
  journal={arXiv preprint arXiv:2608.02365},
  year={2026}
}

@article{li2026ace,
  title={ACE-Ego-0: Unifying Egocentric Human and Robotic Data for VLA Pretraining},
  author={Li, Hao and Zhao, Ganlong and Liu, Yufei and Hou, Haotian and Ye, Guoquan and Fang, Tongyan and Liu, Chunxiao and Huang, Siyuan and Liu, Jianbo and Wang, Xiaogang and others},
  journal={arXiv preprint arXiv:2606.17200},
  year={2026}
}

@article{zhang2026joyai,
  title={Joyai-ra 0.1: A foundation model for robotic autonomy},
  author={Zhang, Tianle and Yuan, Zhihao and Chi, Dafeng and Liu, Peidong and Li, Dongwei and Hu, Kejun and Zhang, Likui and Nie, Junnan and Wei, Ziming and Chen, Zengjue and others},
  journal={arXiv preprint arXiv:2604.20100},
  year={2026}
}

@article{luo2026being,
  title={Being-h0. 7: A latent world-action model from egocentric videos},
  author={Luo, Hao and Zhang, Wanpeng and Feng, Yicheng and Zheng, Sipeng and Xu, Haiweng and Xu, Chaoyi and Xi, Ziheng and Fu, Yuhui and Lu, Zongqing},
  journal={arXiv preprint arXiv:2605.00078},
  year={2026}
}

@article{yang2026abot,
  title={Abot-m0: Vla foundation model for robotic manipulation with action manifold learning},
  author={Yang, Yandan and Zeng, Shuang and Lin, Tong and Chang, Xinyuan and Qi, Dekang and Xiao, Junjin and Liu, Haoyun and Chen, Ronghan and Chen, Yuzhi and Huo, Dongjie and others},
  journal={arXiv preprint arXiv:2602.11236},
  year={2026}
}

@article{yang2026beyond,
  title={Beyond Data Scaling: Representation-Centric Continued Pre-training for Vision-Language-Action Models},
  author={Yang, Senqiao and Wang, Chengyao and Chen, Yuxin and Wang, Zixuan and Tang, Longxiang and Gui, Haokun and Ye, Jinhui and Lu, Changsheng and Wu, Xiaoyang and Zhu, Mingkang and others},
  journal={arXiv preprint arXiv:2608.27550},
  year={2026}
}

@article{xie2026turbovla,
  title={TurboVLA: Real-time vision-language-action model at 32 Hz on an RTX 4090 with< 1 GB VRAM},
  author={Xie, Hengyi and Yao, Chenfei and Wu, Xianjin and Zhu, Yingying and Liang, Dingkang and Bai, Xiang and Ding, Han},
  journal={arXiv preprint arXiv:2607.27205},
  year={2026}
}

@article{li2026light,
  title={Light-WAM: Efficient World Action Models with State-Fusion Action Decoding},
  author={Li, Ziang and Cheng, Dongzhou and Wang, Yibin and Wang, Shiyue and Xu, Xiaoyang and Weng, Lingxuan and Wang, Juan and Wang, Jiaqi},
  journal={arXiv preprint arXiv:2606.08242},
  year={2026}
}

@article{guo2026unified,
  title={Unified 4d world action modeling from video priors with asynchronous denoising},
  author={Guo, Jun and Li, Qiwei and Li, Peiyan and Chen, Zilong and Sun, Nan and Su, Yifei and Wang, Heyun and Zhang, Yuan and Li, Xinghang and Liu, Huaping},
  journal={arXiv preprint arXiv:2604.26694},
  year={2026}
}

@article{community2026starvla,
  title={StarVLA: A Lego-like Codebase for Vision-Language-Action Model Developing},
  author={Community, StarVLA},
  journal={arXiv preprint arXiv:2604.05014},
  year={2026}
}

@article{li2026efficient,
  title={Efficient-wam: A 1b-parameter world-action model with low-cost future imagination},
  author={Li, Jiajun and Guo, Tiecheng and Ye, Yifan and Zhang, Rongyu and Chi, Xiaowei and Sun, Qianpu and Li, Ying and Lou, Yunfan and Wang, Jinqiao and Lu, Zhihe and others},
  journal={arXiv preprint arXiv:2606.10040},
  year={2026}
}

@article{wu2026pragmatic,
  title={A pragmatic vla foundation model},
  author={Wu, Wei and Lu, Fan and Wang, Yunnan and Yang, Shuai and Liu, Shi and Wang, Fangjing and Zhu, Qian and Sun, He and Wang, Yong and Ma, Shuailei and others},
  journal={arXiv preprint arXiv:2601.18692},
  year={2026}
}

@inproceedings{chi2023diffusion,
  title={{Diffusion Policy}: Visuomotor Policy Learning via Action Diffusion},
  author={Chi, Cheng and Feng, Siyuan and Du, Yilun and Xu, Zhenjia and Cousineau, Eric and Burchfiel, Benjamin CM and Song, Shuran},
  booktitle={Proceedings of Robotics: Science and Systems},
  year={2023},
  address={Daegu, Republic of Korea},
  month={July},
  doi={10.15607/RSS.2023.XIX.026}
}

@article{intelligence2025pi05,
  title={\(\pi_{0.5}\): A Vision-Language-Action Model with Open-World Generalization},
  author={{Physical Intelligence} and Black, Kevin and Brown, Noah and Darpinian, James and Dhabalia, Karan and Driess, Danny and Esmail, Adnan and Equi, Michael and Finn, Chelsea and Fusai, Niccolo and Galliker, Manuel Y. and Ghosh, Dibya and Groom, Lachy and Hausman, Karol and Ichter, Brian and Jakubczak, Szymon and Jones, Tim and Ke, Liyiming and LeBlanc, Devin and Levine, Sergey and Li-Bell, Adrian and Mothukuri, Mohith and Nair, Suraj and Pertsch, Karl and Ren, Allen Z. and Shi, Lucy Xiaoyang and Smith, Laura and Springenberg, Jost Tobias and Stachowicz, Kyle and Tanner, James and Vuong, Quan and Walke, Homer and Walling, Anna and Wang, Haohuan and Yu, Lili and Zhilinsky, Ury},
  journal={arXiv preprint arXiv:2504.16054},
  year={2025}
}

@misc{easywam2026,
  title={{EasyWAM}: A Unified and Efficient Framework for Training and Evaluating World Action Models},
  author={{EasyWAM-Team}},
  year={2026},
  howpublished={\url{https://github.com/OpenMOSS/EasyWAM}},
  note={Accessed November 17, 2026}
}

@article{zheng2025xvla,
  title={{X-VLA}: Soft-Prompted Transformer as Scalable Cross-Embodiment Vision-Language-Action Model},
  author={Zheng, Jinliang and Li, Jianxiong and Wang, Zhihao and Liu, Dongxiu and Kang, Xirui and Feng, Yuchun and Zheng, Yinan and Zou, Jiayin and Chen, Yilun and Zeng, Jia and Zhang, Ya-Qin and Pang, Jiangmiao and Liu, Jingjing and Wang, Tai and Zhan, Xianyuan},
  journal={arXiv preprint arXiv:2510.10274},
  year={2025}
}


\end{document}